\documentclass{article}
\usepackage{spconf,amsmath,graphicx}
\usepackage{array}
\usepackage{times}
\usepackage{epsfig}
\usepackage{graphicx}
\usepackage{amssymb}
\usepackage{amsbsy}
\usepackage{float}

\usepackage[subtle]{savetrees}
\usepackage{tabularx}
\newcolumntype{Y}{>{\centering\arraybackslash}X}

\title{Cross-Modal Information Maximization for Medical Imaging: CMIM}
\name{\begin{tabular}{c}Tristan Sylvain$^{\star}$, Francis Dutil$^{\dagger}$, Tess Berthier$^{\dagger}$, Lisa Di Jorio$^{\dagger}$, \\Margaux Luck$^{\star}$, Devon Hjelm $^{\ddagger}$, Yoshua Bengio$^{\star}$\end{tabular}}

\address{$^{\star}$ Mila, Universit\'e de Montreal \\
  $^{\dagger}$Imagia Cybernetics \\
  $^{\ddagger}$Microsoft Research}

\begin{document}
%
© 2020 IEEE. Personal use of this material is permitted. Permission from IEEE must be obtained for all
other uses, in any current or future media, including reprinting/republishing this material for advertising
or promotional purposes, creating new collective works, for resale or redistribution to servers or lists, or
reuse of any copyrighted component of this work in other works.

\maketitle
\begin{abstract}
In hospitals, data are siloed to specific information systems that make the same information available under different modalities such as the different medical imaging exams the patient undergoes (CT scans, MRI, PET, Ultrasound, etc.) and their associated radiology reports. This offers unique opportunities to obtain and use at train-time those multiple views of the same information that might not always be available at test-time.

In this paper, we propose an innovative framework that makes the most of available data by learning good representations of a multi-modal input that are resilient to modality dropping at test-time, using recent advances in mutual information maximization. By maximizing cross-modal information at train time, we are able to outperform several state-of-the-art baselines in two different settings, medical image classification, and segmentation. In particular, our method is shown to have a strong impact on the inference-time performance of weaker modalities.
\end{abstract}

\section{Introduction}
\label{sec:intro}
The practice of keeping hospital patient data inside information silos restricts the range of possible data analyses that could improve patient care. The richness of hospital databases that manifests itself in their increasing volumes and modalities/sources could offer unique opportunities for data analysis improvement through the acquisition and use of multiple views of the same patient coming from, for example, different medical imaging exams (CT scans, MRI, PET, Ultrasound) and associated radiology reports.

However, patient data may contain a large variety of modalities, many of which may be missing for a specific patient due to differing clinical procedures between specialists and hospitals. This occurs when we are considering a specific modality in medical data sets that tends to only be present in a few data points (for instance due to cost or rarity of the medical condition requiring it). In addition, for some modalities such as radiology reports and imagery, hospital-specific guidelines can lead to non-standardized annotations, image acquisition artifacts, etc. This leads to medical data sets that are often too sample-poor to fully take advantage of deep learning techniques. A solution to this problem would be to build a deep learning model that will take advantage of the multiple modalities available at training time by learning single-modal representations that minimize the information loss when compared to multi-modal representations of the same input. This would encourage robustness to modality dropping (i.e., the model must be able to perform well in the absence of one or more modalities) at testing time. A way to do that is to apply recent advances in mutual information maximization~\cite{hjelm2018learning,belghazi2018mine}.

In this paper, we idealize this problem setting by considering extreme modality dropping at testing time (i.e., multiple modalities at train time, one at test time) to improve classification of chest x-rays using the open-source Open-I data set and the segmentation of different MRI modalities using the publicly available BRATS-2015 data set. Our contributions are as follows:
\begin{itemize}
    \item We reformulate cross-modal training as a mutual information maximization problem, and propose an innovative framework harnessing recent advances in mutual-information estimation to address it.
    \item By design, we are able to exploit learned representations for every modality and exploit them at test time even when one modality is missing.
    \item Our proposed approach outperforms state-of-the-art baselines on two challenging tasks, image classification and semantic segmentation.
\end{itemize}

\section{Related Work}\label{sec:relatedwork}
\subsection{Cross-modality training}
Multi-modal data has been exploited in numerous medical tasks including: caption generation~\cite{wang2018tienet} (text and images), lesion detection~\cite{hadad2017classification} (mammogram and MRI), image classification~\cite{zhang2019knowledge} (image and knowledge graphs) and few-shot semantic segmentation~\cite{Zhao_2019_CVPR}. While such systems yield performance improvements, there are few works on creating systems that while benefiting from additional training modalities are \emph{robust} to modality dropping at test-time.

Solutions generally fall into three broad categories. In the first case, missing modalities are inferred at test-time via e.g. retraining a model with the missing modalities~\cite{hofmann2008mri}, synthesizing missing-modalities~\cite{van2015does}, or bootstrapping from a classifier trained on the full set of features~\cite{hor2015scandent}. The second approach maps modalities to a common subspace via e.g. an abstraction layer focusing on first-order statistics~\cite{havaei2016hemis} or adversarial methods~\cite{saito2016demian,sylvain2020object}. The third, to which our method belongs, optimizes some similarity metric between different views/modalities of the data, by e.g. canonical correlation analysis~\cite{hotelling1992relations, andrew2013deep} or attention combined with shared tasks such as MDNet~\cite{zhang2017mdnet} and TieNet~\cite{wang2018tienet}.

\subsection{Mutual information maximization}
Mutual information (MI), despite being a useful quality to evaluate, is hard to estimate in practice for non-discrete representations. Mutual Information Neural Estimation \cite{belghazi2018mine} introduces an estimator of mutual information via an auxiliary network. Deep InfoMax \cite{hjelm2018learning} and more recently AM-DIM~\cite{bachman2019learning} apply this framework to representation tasks by maximizing mutual information between local and global representations of an input. ST-DIM~\cite{anand2019unsupervised} and CM-DIM~\cite{sylvain2020locality,sylvain2020zero} apply this in turn to reinforcement learning and zero-shot learning respectively. Our work is the first to consider maximizing mutual information between representations of different modalities of a same input.

Previous works on applying mutual information to cross-modal learning usually constrain the architecture, such as the shared weights approach of \cite{rastegar2016mdl} or introduce other constraints whereas our approach is more general.

\section{Proposed method}
\begin{figure*}
  \centering
    \includegraphics[trim=0cm 5mm 0cm 0cm,width=\textwidth, height=7cm]{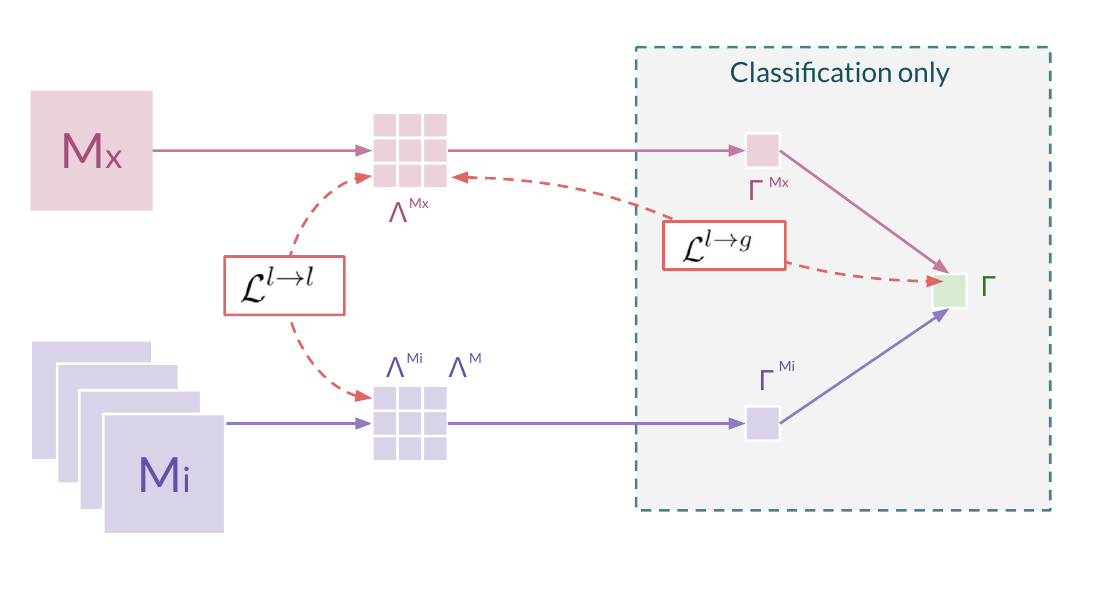}
  \caption{Graphical representation of our method on multiple modalities. We train on a set $\mathbf{M} = \big\{\mathbf{M_X}, \mathbf{M_1}, \cdots, \mathbf{M_n} \big\}$ modalities at train time. At test-time, only $\mathbf{M_X}$ is provided. We map the input to both per-modality local features $\mathbf{\Lambda^{M_i}}$, per-modality global features $\mathbf{\Gamma^{M_i}}$ and a multi-modal global embedding $\mathbf{\Gamma}$ shared across modalities. The local-local and local-global losses correspond to the mutual information terms introduced in the proposed method section. We omit the global embeddings for the segmentation task. In addition to the losses shown, we also train with the task specific segmentation loss $\mathcal{L}^{\text{seg}}$ (pixel-wise categorical cross-entropy) and classification loss $\mathcal{L}^{\text{classif}}$ (categorical cross entropy) not shown in this figure.} 
  \label{fig:model}
\end{figure*}
Our approach, represented in Figure~\ref{fig:model}, aims to improve supervised downstream performance in the setting where a subset of modalities present at train time are not present at test time\footnote{In practice we consider only one modality present at test-time}. We do this by maximizing mutual-information between representations of different modalities of a given input. This will encourage each modality to retain as much discriminative information as possible.

\subsection{Mutual information maximization}
Our work applies the mutual information neural estimator (MINE) introduced in~\cite{belghazi2018mine}. Formally, the mutual information between two random variables $X$ and $Y$ is defined as the KL-divergence between the joint distribution and the product of the marginals, i.e., $\mathcal{D}_{KL}(\mathbb{P}_{XY} || \mathbb{P}_X \otimes \mathbb{P}_Y)$. MINE maximizes a lower bound on that quantity derived from the Donsker Varadhan formulation. In our case, we found, similarly to~\cite{hjelm2018learning}, that performance was improved by considering instead the Jensen-Shanon estimator, leading to: 
\begin{align*}
    \hat{I}_\theta(X, Y) = \mathbb{E}_{\mathbb{P}_{XY}}[-\text{sp}(T_\theta(x, y))] -\mathbb{E}_{\mathbb{P}_X \otimes \mathbb{P}_Y}[\text{sp}(T_\theta(x, y)],
\end{align*}
where $\text{sp}(z) = \log(1+e^z)$ and $T_\theta$ is a neural network with parameters $\theta$.

\subsection{Cross-modality mutual information}
In this work, we are concerned with optimizing mutual information between representations of different modalities of a given input. We train on a set $\mathbf{M} = \big\{\mathbf{M_X}, \mathbf{M_1}, \cdots, \mathbf{M_n} \big\}$ modalities at train time. At test-time, only $\mathbf{M_X}$ is provided. Each modality $\mathbf{M_i}$ can be mapped to local features $\mathbf{\Lambda^{M_i}}$ (2D and 1D pre-pooling convolution maps for images, and text respectively), and global features (pooled convolution maps) $\mathbf{\Gamma^{M_i}}$. Similarly, we can obtain representations $\mathbf{\Lambda^{M}}$ and $\mathbf{\Gamma^{M}}$ for all the input modalities. For more details, see the section on design choices.
We can then define cross-modal local-local, local-global and global-global losses as respectively:
\begin{align*}
    &\mathcal{L}^{l\rightarrow l} = \frac{1}{N^2}\sum_{n, m} ^N \hat{I}(\mathbf{\Lambda^{M_i}}_n, \mathbf{\Lambda^{M}}_m) \\
    &\mathcal{L}^{l\rightarrow g} = \frac{1}{N}\sum_n \hat{I}(\mathbf{\Lambda^{M_i}}_n, \mathbf{\Gamma^{M}}) \\ 
    &\mathcal{L}^{g\rightarrow g} = \hat{I}(\mathbf{\Gamma^{M_i}}, \mathbf{\Gamma^{M}})
\end{align*}

\subsection{Design choices}
We have presented a global framework that can tackle different cases. In what follows, we will apply it to \emph{semantic segmentation} and \emph{image classification}. We only optimize the local-local mutual information loss in the first case. For classification, we optimize two losses: local-local and local-global. This choice is motivated by empirical performance, and the argument that semantic segmentation tasks benefit less from global information.


Each model is in addition to the mutual information losses optimized with its task-specific loss, $\mathcal{L}^{\text{seg}}$ (pixel-wise categorical cross-entropy) and $\mathcal{L}^{\text{classif}}$ (categorical cross-entropy).

The final training classification and segmentation losses are respectively:
\begin{align*}
    &\mathcal{L}^C = \lambda_{l\rightarrow g} \mathcal{L}^{l\rightarrow g} + \lambda_{l\rightarrow l} \mathcal{L}^{l\rightarrow l} + \lambda_C \mathcal{L}^{\text{classif}}\\
    &\mathcal{L}^S = \lambda_{l\rightarrow l} \mathcal{L}^{l\rightarrow l} + \lambda_S \mathcal{L}^{\text{seg}}\\
\end{align*}
where the $\lambda$ are hyper-parameters regulating the importance of the different losses during training.

We use different architectures for the two downstream classification and segmentation tasks. For the segmentation task, we consider 4 MR \emph{modalities} (FLAIR, T1W, T1C, T2), which are encoded using a U-Net~\cite{ronneberger2015u}-type model, due to its use in past literature, and overall good performance in medical segmentation. It takes as input either one or multiple MRI sequences as distinct channels to compute the representations. 

For the classification setting, two \emph{modalities} are present: text and image. Text was encoded using 300-dimension Glove vectors~\cite{pennington2014glove} trained on Wikipedia. We did not perform fine-tuning of the embedding as this negatively impacted performance. Image representations are obtained using a ResNet50~\cite{he2016deep} encoder, and text representations using a residual CNN variant of~\cite{zhang2017deconvolutional}. The global embedding is a bilinear embedding of the two previous representations.

For $T_\theta$, we used architectures similar to the ``concat-and-convolve" architecture found in \cite{hjelm2018learning} (see Figure 5 in \cite{hjelm2018learning}).

\section{Experiments}
\begin{table}
\centering
\begin{tabularx}{\columnwidth}{@{}|Y|Y|Y|Y|@{}}
\hline
Method & \begin{tabular}[t]{@{}c c c@{}} 
         \phantom{h}Training\phantom{h}\\
          \thinspace \thinspace \thinspace I \thinspace  \thinspace  \thinspace  \thinspace \thinspace  T\\
      \end{tabular} &  \begin{tabular}[t]{@{}c c c@{}} 
         \phantom{h}Testing\phantom{h}\\
          \thinspace \thinspace \thinspace I \thinspace  \thinspace  \thinspace  \thinspace \thinspace T*\\
      \end{tabular}
      & \phantom{h}AUC\phantom{h} \\
\hline \hline
      ResNet &
      \begin{tabular}[t]{@{}ccc@{}}\phantom{h}$\bullet$ & $\circ$ \end{tabular} &
    \begin{tabular}[t]{@{}ccc@{}} $\bullet$ & $\circ$ \end{tabular} & $0.785$\\
    TieNet & 
      \begin{tabular}[t]{@{}ccc@{}}\phantom{h}$\bullet$ & $\bullet$ \end{tabular} &
    \begin{tabular}[t]{@{}ccc@{}} $\bullet$ & $\bullet$ \end{tabular} & 0.741\\
   \hline \hline
    CMIM & 
      \begin{tabular}[t]{@{}ccc@{}}\phantom{h}$\bullet$ & $\bullet$ \end{tabular} &
    \begin{tabular}[t]{@{}ccc@{}} $\bullet$ & $\circ$ \end{tabular} & $\mathbf{0.793}$\\
\hline
\end{tabularx}
\caption{Results on Open-i. The \textit{Train phase} and \textit{Test phase} columns indicates which modality were used, among the image \textbf{I}, the text \textbf{T}, and the generated text \textbf{T*} obtained from a captioning model ($\bullet$ denotes presence, $\circ$ absence). Note that the true text modality \textbf{T} is never present at test time. As we can see, our model outperforms the baselines, and contrary to TieNet, is actually able to leverage the second modality during training.}
\label{caption_gen_results}
\end{table}

\begin{table*}[ht]
\resizebox{\textwidth}{!}{%
\begin{tabular}{|c|c|c|c|c|}
\hline
\begin{tabular}[t]{@{}c c c c@{}} 
         \phantom{h}Test-time modalities\phantom{h}\\
          \thinspace \thinspace \thinspace \thinspace $F$ \thinspace  \thinspace  \thinspace  \thinspace \thinspace $T_1$ \thinspace  \thinspace  \thinspace  \thinspace \thinspace $T_1c$ \thinspace  \thinspace  \thinspace  \thinspace \thinspace$T_2$\\
      \end{tabular} &\phantom{h}CMIM\phantom{h} & \phantom{h}*HeMIS\phantom{h} & \phantom{h}*Mean (baseline)\phantom{h} & \phantom{h}*MLP (baseline)\phantom{h} \\
\hline \hline
      \begin{tabular}[t]{@{}c c c c@{}}$\bullet$\phantom{.} & $\circ$\phantom{h} & $\circ$\phantom{h} & $\circ$\end{tabular} &
    $\mathbf{23.37}$ & $5.57$ & $6.25$ & $15.90$\\
    
    \begin{tabular}[t]{@{}c c c c@{}}\phantom{h}$\circ$\phantom{.} & $\bullet$\phantom{h} & $\circ$\phantom{h} & $\circ$\phantom{h}\end{tabular} &
    $\mathbf{14.15}$ & $4.67$ & $6.25$ & $10.78$\\
    
    \begin{tabular}[t]{@{}c c c c@{}}\phantom{h}$\circ$\phantom{.} & $\circ$\phantom{h} & $\bullet$\phantom{h} & $\circ$\phantom{h}\end{tabular} &
    $49.00$ & $\mathbf{49.93}$ & $30.02$ & $32.92$\\
    
    \begin{tabular}[t]{@{}c c c c@{}}\phantom{h}$\circ$\phantom{.} & $\circ$\phantom{h} & $\circ$\phantom{h} & $\bullet$\phantom{h}\end{tabular} &
    $\mathbf{29.56}$ & $20.31$ & $6.25$ & $18.62$\\

\hline
\end{tabular}
}
\caption{Dice similarity coefficient (DSC) results on the BRATS test sets (\%) in the "enhancing" setting introduced in~\cite{havaei2016hemis}. We consider the case where only one of the 4 modalities is present at test-time ($\bullet$ denotes presence, $\circ$ absence). All 4 modalities are used at train-time. Note that both these conditions create a very challenging setting, explaining the overall low dice scores reported. * denotes results taken from~\cite{havaei2016hemis}. Our approach outperforms HEMIS and the other baselines on this setting with the exception of the $T_1c$ modality where our model is a close second. In particular, strong gains are observed for "weaker" modalities such as F and $T_1$}
\label{fig:seg_results}
\end{table*}

\subsection{Experimental setup}
For each task, we train using the full set of available modalities, and evaluate using a single modality. Such as setting occurs frequently in practice as per instance there might be a small overlap between the MRI modalities a model has been trained on and the set of acceptable test-time modalities (due to the absence of some, or domain shifts due to device calibration making some modalities unusable)
\subsection{Classification task}
\emph{Open-I}~\cite{demner2015preparing} is a publicly available radiography dataset collected by Indiana University. It contains 7470 chest x-rays with 3955 radiology reports. We prepared the data using the same methodology as~\cite{wang2018tienet}, i.e. keeping 14 categories of findings as the classes for the classification problem, and only considering frontal images with associated reports. As the orientations of the X-ray images are not specified, and in order to keep only the frontal views, we performed manual analysis of all images, also removing some that were heavily distorted. We re-balanced the dataset as the raw data had heavy class imbalance. We report Area under the Curve (AUC) for all methods.

\subsection{Semantic segmentation task}
\emph{BRATS-2015} \cite{menze2015multimodal,bakas2017advancing} is a brain MRI dataset containing 220 subjects with high grade tumors, and 54 subjects with low grade tumors. There are 4 MR \emph{modalities} present (FLAIR, T1W, T1C, T2), alongside a voxel-level segmentation ground truth of 5 labels: \emph{health}, \emph{necrosis}, \emph{edema}, \emph{non-enhancing tumor} and \emph{enhancing tumor}. As in the \emph{enhancing} setting in~\cite{havaei2016hemis}, the target is a binary map corresponding to a 1-versus-rest segmentation on the \emph{enhancing tumor} class.

\subsection{Baselines}
For the classification task, we compare our results to TieNet~\cite{wang2018tienet}, a state of the art method for multi-modal X-ray classification. We also benchmark against a ResNet50~\cite{he2016deep} supervised on the image modality only.
For the segmentation task, we compare with Hemis~\cite{havaei2016hemis}, a state-of-the art approach on this dataset. We also considered the same baselines that Hemis suggested: missing modality completion by mean (Mean) and a multi-layer perceptron (MLP). To ensure conformity with their experimental setup, we used the same splits and code for data preparation.

\subsection{Implementation and Training Details}
\label{sec:implementation}
Our code is written in PyTorch. Each experiment ran on V100 GPUs, using the Adam~\cite{kingma2014adam} solver with a global learning rate of 0.0001. Models were trained up to convergence (early stopping on a validation set).

\subsection{Results}
When applying our model to the two experimental tasks, we had to make small adaptations. As local information tends to be more important in segmentation, we empirically found that local-global and global-global did not improve performance. This was not the case for classification, where we also used the local-global loss.

For the classification setting, as we can see in table \ref{caption_gen_results}, our method outperformed the other baselines. The discrepancy between TieNet's result and the other methods can be explained by the low number of training examples for the captioning model. Indeed, compared to TieNet's original paper where 100 000 reports are available, only a few thousands are present in Open-I. This causes the model to quickly disregard the image modality and to only focus on a few keywords to make its decision.  However, CMIM alleviates this problem by forcing the representation of both modalities to have high mutual-information. This in turn encourages discriminative information to be present in both representations at inference time.

The results for the semantic segmentation task can be seen in table~\ref{fig:seg_results}. Our model outperforms the other methods overall. Interestingly, our approach seems to perform better for the ``weaker'' modalities (F and T1 are known to perform poorly for enhanced tumor detection~\cite{havaei2016hemis}), where less information is present at test-time. This validates our hypothesis that CMIM is able to enhance discriminative features, even when the modality contains a low amount of signal.

\section{Conclusion}
In this paper, we introduced a method based on mutual information for cross-modal training. These kind of approaches can be particularly useful when some modalities are missing, as is often the case with real world data, in particular medical data. We validated our approach in two different tasks, each one implying different type of modality: text and image for a classification task, and different MRI modalities for a segmentation task. In both cases, results are promising. Interestingly, for MRI segmentation, our approach yields the best results when the modality present at test time conveys less discriminative information.

For future work, we plan on adapting the current model to be able to use multiple modalities at test time. Furthermore, we hope that our setup will pave the way for zero-shot learning approaches, where we would present the model with unseen modalities at test time.

\bibliographystyle{IEEEbib}
\bibliography{bibfile}

\end{document}